\documentclass[runningheads,orivec]{llncs}
\usepackage[T1]{fontenc}
\usepackage{graphicx,verbatim}
\usepackage{amsmath, amssymb}
\usepackage{xcolor}
\usepackage[normalem]{ulem}

\begin{document}
\title{Breast Cancer Detection from Multi-View Screening Mammograms with Visual Prompt Tuning}

%
\author{Han Chen\inst{1,2} \and
Anne L. Martel\inst{1,2}}
\institute{Physical Sciences, Sunnybrook Research Institute, Toronto, ON, Canada \and
Medical Biophysics, University of Toronto, Toronto, ON, Canada}


\maketitle              
\begin{abstract}

Accurate detection of breast cancer from high-resolution mammograms is crucial for early diagnosis and effective treatment planning. Previous studies have shown the potential of using single-view mammograms for breast cancer detection. However, incorporating multi-view data can provide more comprehensive insights. Multi-view classification, especially in medical imaging, presents unique challenges, particularly when dealing with large-scale, high-resolution data. In this work, we propose a novel Multi-view Visual Prompt Tuning Network (MVPT-NET) for analyzing multiple screening mammograms. We first pretrain a robust single-view classification model on high-resolution mammograms and then innovatively adapt multi-view feature learning into a task-specific prompt tuning process. This technique selectively tunes a minimal set of trainable parameters (7\%) while retaining the robustness of the pre-trained single-view model, enabling efficient integration of multi-view data without the need for aggressive downsampling. Our approach offers an efficient alternative to traditional feature fusion methods, providing a more robust, scalable, and efficient solution for high-resolution mammogram analysis. Experimental results on a large multi-institution dataset demonstrate that our method outperforms conventional approaches while maintaining detection efficiency, achieving an AUROC of 0.852 for distinguishing between Benign, DCIS, and Invasive classes. This work highlights the potential of MVPT-NET for medical imaging tasks and provides a scalable solution for integrating multi-view data in breast cancer detection.



\end{abstract}

\section{Introduction}
Breast cancer is a leading cause of cancer-related mortality among women worldwide, representing a significant public health challenge \cite{contiero2023causes}. Early detection through screening mammography is essential for improving patient outcomes and reducing mortality rates. Mammography remains the most reliable tool for early breast cancer diagnosis, typically capturing two high-resolution images of each breast from different perspectives: mediolateral-oblique (MLO) and craniocaudal (CC). Unlike conventional deep learning approaches that primarily rely on single-view analysis, radiologists interpret all views together, drawing on cross-view correlations to improve diagnostic accuracy. This practice illustrates the value of multi-view analysis in identifying abnormalities and supporting clinical decision-making. As a result, multi-view or multi-image-based Computer-Aided Diagnosis (CAD) systems have gained attention for their potential to surpass the performance of single-view models \cite{chen2022multi,black2024multi}. Recent advancements in deep learning have further enhanced breast cancer classification and detection, with increasing interest in multi-view architectures inspired by radiologists’ interpretive strategies \cite{jain2024follow}. These developments suggest that multi-view learning could play a vital role in advancing breast cancer detection and diagnosis.

Recent advances in artificial intelligence, particularly deep neural networks, have demonstrated remarkable capabilities in extracting informative imaging features for CAD \cite{gou2024research,chen2023teacher,chen2024towards}. Among various deep learning architectures \cite{vaswani2017attention,chen2022unsupervised,krizhevsky2017imagenet,liu2021swin}, Transformers \cite{vaswani2017attention} have gained significant attention in medical image analysis. Transformers partition an image into a sequence of small patches and employ multi-head self-attention mechanisms to model dense long-range dependencies. Their self-attention design effectively captures both local and global features, providing an advantage over Convolutional Neural Networks (CNNs) \cite{krizhevsky2017imagenet}. Despite the success, Transformers suffer from quadratic time complexity relative to sequence length, making them computationally expensive for high-resolution images. This issue is particularly critical in multi-view mammogram analysis, where leveraging correlations across different views can significantly enhance diagnostic accuracy. Although several methods \cite{sarker2024mv,van2021multi,li2020multi} have been proposed for multi-view mammogram analysis, their performance is often constrained by aggressive downsampling. This downsampling results in a loss of fine-grained details, which are crucial for accurate breast cancer detection. Consequently, current models struggle to fully exploit the rich information contained in high-resolution, multi-view mammograms.

Recently, Visual Prompt Tuning (VPT) \cite{jia2022visual} has emerged as an efficient method for adapting large pre-trained vision Transformers to new tasks. Originating from prompting \cite{sahoo2024systematic} in natural language processing, VPT extends the concept of prepending task-specific instructions to input data, enabling models to generalize to new tasks with minimal adjustments. Unlike traditional fine-tuning, which updates all model parameters, VPT introduces a small set of learnable prompts in the input space while keeping the backbone frozen, significantly reducing the number of trainable parameters. Motivated by its efficiency and strong performance in vision tasks, recent works have increasingly applied VPT to medical image processing \cite{he2025dvpt,zu2024embedded}.

In this study, we propose a novel Multi-view Visual Prompt Tuning Network (MVPT-NET) for efficient multi-view mammogram classification in breast cancer detection. We treat this task as a two-stage learning process. Firstly, to leverage the power of pre-trained models, we pretrain a Swin Transformer on a single-view classification task. Secondly, during downstream prompt tuning, a small set of task-specific learnable parameters from multi-view inputs is introduced into the input space, while the entire pre-trained Swin Transformer backbone remains frozen. To further enhance the cross-view correlation learning, tokens from different views are merged and passed through the transformer blocks to obtain the final comprehensive prediction. MVPT-NET sets itself apart from conventional approaches by efficiently utilizing pre-trained single-view models, and tuning with multi-view data, retaining high-resolution information, and avoiding aggressive downsampling. We use data from 4,179 subjects across multiple NHS Breast Cancer screening sites in the UK \cite{halling2020optimam} to evaluate the performance of MVPT-NET for breast cancer classification. Experimental results show that MVPT-NET achieves an Area under the Receiver Operating Characteristic Curve (AUROC) of 0.852 for distinguishing between Benign (Normal), DCIS, and Invasive cancers. Notably, only 7\% of all model parameters are tuned during training. Our method demonstrates strong performance with a relatively small percentage of parameter adjustments, highlighting its potential for multi-view mammogram analysis in breast cancer detection.

\section{Methods}
In this section, we describe our baseline model and MVPT-NET (illustrated in Fig.\ref{fig1}). Our proposed method involves two phases of training: single-view baseline model pretraining and downstream multi-view visual prompt tuning with mammogram pairs, along with cross-view knowledge distillation. 

\subsection{Single-view Pretraining}
Our single-view baseline model follows the Swin Transformer Base architecture \cite{liu2021swin}, which is chosen as the backbone for MVPT-NET due to its hierarchical feature representation and efficient modeling of long-range dependencies using shifted window attention. Given an input mammogram \( I \in \mathbb{R}^{H \times W \times C} \), where \( H \) and \( W \) are the height and width of the input image, and \( C \) is the number of channels, the image is first divided into fixed-sized patches of size \( h \times w \). These patches are then linearly embedded into a latent space, producing patch tokens \( \text{E}_0 \in \mathbb{R}^{m \times d} \), where \( m = \frac{H}{h} \times \frac{W}{w} \) denotes the number of patches, and \( d \) is the embedding dimension:

\begin{figure}[t]
\includegraphics[width=\textwidth]{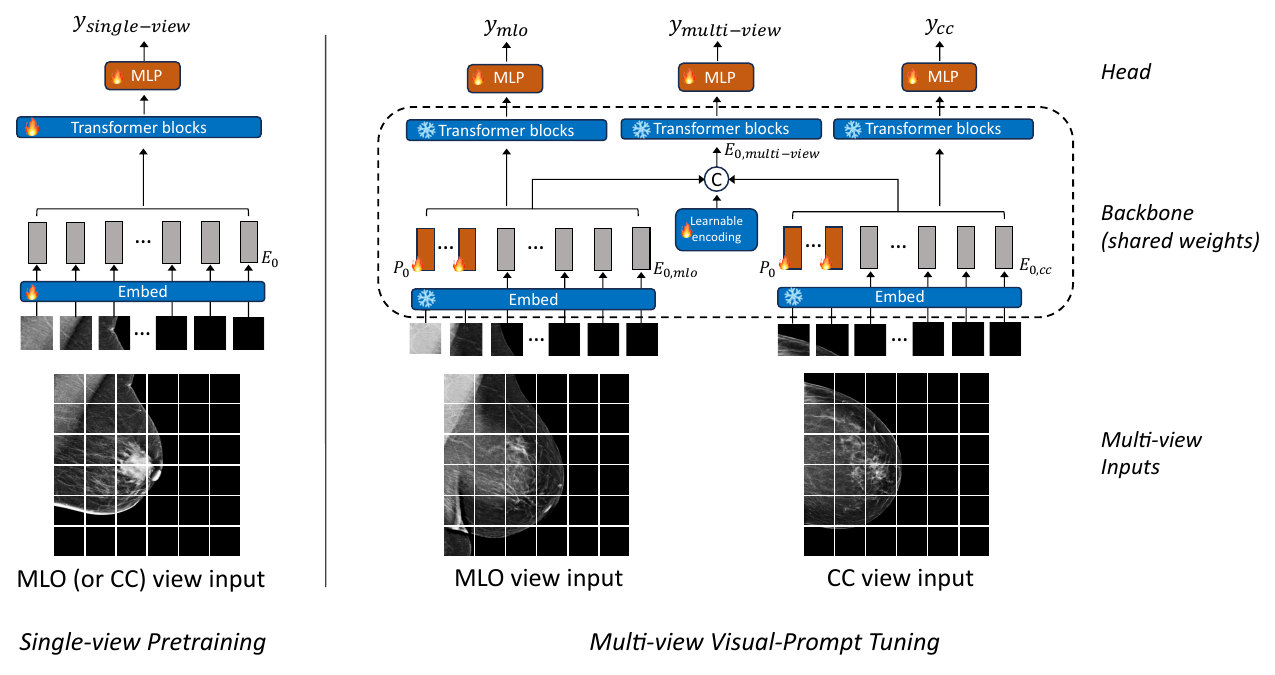}
\caption{Network architecture of MVPT-NET.} 
\label{fig1}
\end{figure}

\begin{equation}
\text{E}_0 = \text{Embed}(I), \quad \text{E}_0 \in \mathbb{R}^{m \times d}
\end{equation}

The embedded patches are then passed through a series of Swin Transformer layers. Each layer \( L_i \) consists of shifted window-based multi-head self-attention (W-MSA) and feed-forward networks (FFN) with residual connections and LayerNorm. The computation in each layer \( L_i \) is formulated as:

\begin{equation}
\text{E}_i = L_i(\text{E}_{i-1}), \quad i = 1, 2, \dots, N
\end{equation}

\noindent where \( N \) is the total number of Swin Transformer layers. Finally, a neural classification head maps the output of the final layer \( \text{E}_N \) into a predicted class probability distribution \( y \):

\begin{equation}
y = \text{Head}(\text{E}_N)
\end{equation}

The \( y \) is then used to calculate the cross-entropy loss with the single-view mammogram ground truth. The pre-trained Swin Transformer model will serve as the backbone for MVPT-NET, which will be further prompt-tuned in the downstream task of multi-view mammogram classification.




\subsection{Multi-view Visual Prompt Tuning}
\subsubsection{Preliminaries.} 
Visual Prompt Tuning (VPT) \cite{jia2022visual} introduces a set of \( p \) continuous embeddings, or prompts, of dimension \( d \) into every Transformer layer’s input space after the embedding layer. During tuning, only the task-specific prompts are updated, while the Transformer backbone remains frozen. For the \( (i+1) \)-th layer \( L_{i+1} \), the collection of learnable prompts is denoted as \( \text{P}_i = \{ \text{p}_{k}^{i} \in \mathbb{R}^d \mid k \in \mathbb{N}, 1 \leq k \leq m \} \). The prompted Swin Transformer model can then be formulated as Equations \eqref{eq:5}, where the symbols \textcolor{red}{\underline{\hbox to 2mm{}}} and  \textcolor{blue}{\uwave{\hbox to 2mm{}}} indicate \textcolor{red}{\underline{learnable}} and  \textcolor{blue}{\uwave{frozen}} parameters, respectively:




\begin{equation}
\begin{aligned}
\text{E}_{i, \text{view}} &= \textcolor{blue}{\uwave{L_i}}([\textcolor{red}{\underline{\text{P}_{i-1}}}, \text{E}_{i-1, \text{view}}]), \quad i = 1, 2, \dots, N \\
y_{\text{view}} &= \textcolor{red}{\underline{\text{Head}}}(\text{E}_N), \quad \text{view} \in \{mlo,cc\}
\end{aligned}
\label{eq:5}
\end{equation}

\subsubsection{Multi-view Visual Prompt Tuning.} Our proposed MVPT-NET adapts VPT into multi-view mammogram classification. Given paired mammograms as input, the patch embedding layer processes the MLO and CC views respectively, generating tokens denoted as \( \text{E}_{0,{mlo}} \) and \( \text{E}_{0,{cc}} \). Previous studies have explored various fusion strategies for integrating multi-view mammograms in breast cancer detection, such as middle fusion and late fusion. Inspired by the work \cite{black2024multi}, we adopt an early fusion strategy to effectively combine tokens from multi-view. Specifically, the joint token representation is formulated as:  


\begin{equation}
\begin{aligned}
\text{E}_{0,{joint}} &= [\textcolor{red}{\underline{\text{P}_0}}, \text{E}_{0,{mlo}}; \textcolor{red}{\underline{\text{P}_0}}, \text{E}_{0,{cc}}] \\
\text{E}_{0, \text{multi-view}} &= \text{E}_{0,{joint}} + \textcolor{red}{\underline{\text{E}_{\text{context}}}}
\end{aligned}
\label{eq:6}
\end{equation}

\noindent where \( \text{E}_{\text{context}} \) is a learnable encoding shared across all tokens from each view.  This enriched multi-view representation is then fed into the Swin Transformer layers as Equation \eqref{eq:7}, enabling the model to capture comprehensive cross-view correlations and get the overall prediction \(y_{\text{multi-view}}\) for accurate mammogram classification.


\begin{equation}
\begin{aligned}
\text{E}_{i, \text{multi-view}} &= \textcolor{blue}{\uwave{L_i}}([\text{E}_{i-1, \text{multi-view}}]), \quad i = 1, 2, \dots, N \\
y_{\text{multi-view}} &= \textcolor{red}{\underline{\text{Head}}}(\text{E}_{N, \text{multi-view}})
\end{aligned}
\label{eq:7}
\end{equation}

\subsubsection{Learning Object and Knowledge Distillation.} The overall training loss is the combination of four terms:

\begin{equation}
L_{\text{overall}} = L_{\text{multi-view}} + L_{\text{mlo}} + L_{\text{cc}} + \lambda L_{\text{md}}
\end{equation}

\noindent where the first three terms represent the cross-entropy loss between \( y_{\text{multi-view}} \), \( y_{\text{mlo}} \), \( y_{\text{cc}} \) and the ground truth label. \( \lambda \) is a trade-off parameter to balance the contribution of the distillation term \( L_{\text{md}} \) and the classification terms. 

Inspired by the work \cite{black2024multi}, we incorporate a mutual knowledge distillation loss term \( L_{\text{md}} \) to encourage collaboration between the \( y_{\text{mlo}} \), \( y_{\text{cc}} \), and \( y_{\text{multi-view}} \). This term is based on the distillation scheme of work \cite{hinton2015distilling}, which minimizes the Kullback-Leibler divergence between temperature-softened distributions produced by the teacher and student models. The loss is calculated as:

\begin{equation}
L_{\text{kd}}(t, s; \tau) = D_{\text{KL}}(\tilde{\sigma}(t, \tau), \tilde{\sigma}(s, \tau))
\end{equation}

The mutual distillation loss \( L_{\text{md}} \) is formulated as a combination of asymmetric distillation terms between the score-fused class distribution and the multi-view predictions:

\begin{equation}
L_{\text{md}}(\{y_{\text{mlo}}, y_\text{cc}\},  y_\text{multi-view}; \tau) = \frac{1}{\tau^2} \left[ L_{\text{kd}}(\hat{\bar{z}}, y_\text{multi-view}; \tau) + L_{\text{kd}}(\hat{y}_{\text{multi-view}}, \bar{z}; \tau) \right]
\end{equation}

\noindent where \( \bar{z} = \frac{1}{2}(y_\text{mlo}+ y_\text{cc})\), and the \( \hat{\bar{z}} \) represents the gradient-detached copy of \( \bar{z} \). This loss term helps align the logits from the multi-view and single-view predictions, improving generalization and performance across different views.

\section{Experiments and Results}
\subsection{Datasets}
This study uses a subset of the OPTIMAM dataset \cite{halling2020optimam} \cite{website}, which includes mammogram images and clinical data from four NHS Breast Cancer screening sites across the UK. All mammograms in the subset were acquired using Hologic or Siemens systems, with a ratio of 2:1. Subjects are categorized into two label groups according to their clinical records: (1) Benign (Normal) vs. Malignant and (2) Benign (Normal) vs. DCIS vs. Invasive. 
In the first group, a subject is classified as benign (normal) if the biopsy confirms a benign (normal) result, and as malignant if the biopsy confirms a malignant result. In the second group, benign (normal) subjects are the same as in the first group, while DCIS subjects have a malignant biopsy result with DCIS grade recorded but no invasive components. Invasive subjects are those with biopsy-confirmed malignant results that include invasive components.

The dataset includes 8,358 mammograms from 4,179 subjects, each with two mammograms (MLO and CC views). It comprises 1,500 benign (normal) and 2,679 malignant subjects in the first group, and 1,500 benign (normal), 1,475 DCIS, and 1,204 invasive subjects in the second group. The data is split into training (80\%, 3,343 subjects) and test (20\%, 836 subjects) sets, with the training set further divided into 5 folds for cross-validation. Category proportions are preserved across all splits.

\subsection{Implementation Details}
We set the input mammogram size to \(1024\times 1024\) and use a Swin Transformer Base \cite{liu2021swin} architecture with an increased window size (from 7 to 16). For single-view pretraining, we initialized the model with self-supervised pretrained weights from ImageNet \cite{russakovsky2015imagenet}, adapting them via bicubic interpolation for the larger window size. The initial learning rate was set to 5e-5 with a batch size of 6. We used the AdamW optimizer with a cosine learning rate scheduler, applying a 10-epoch warm-up starting from 1e-6. The model was trained for 50 epochs.

For multi-view visual prompt tuning, we set the base learning rate to 0.01, weight decay to 0.01, and batch size to 2, using the SGD optimizer with a cosine scheduler and a 10-epoch warm-up. The total number of epochs is 100, and the prompt length is set to 5. For the hyperparameters in \( L_{\text{md}} \), we follow the original setup from \cite{black2024multi}, setting \( \tau = 4 \) and \( \lambda = 0.1 \).

Both pretraining and prompt tuning involved data augmentation, including random vertical flipping and affine transformations. Mammograms were reoriented to ensure the breast consistently appears on the left side. Models were trained on an NVIDIA A100 GPU. The evaluation metrics include accuracy, precision, recall, F1-score, and the area under the receiver operating characteristic (AUROC).

\subsection{Results of Breast Cancer Detection}
We compared our proposed method with several state-of-the-art multi-view classification models, including Multi-View Swin Transformer (MV-Swin-T) \cite{sarker2024mv}, Multi-View Chest Radiograph Classification Network (MVC-NET) \cite{zhu2021mvc}, Cross-View Transformers (CVT) \cite{van2021multi}, and  Multi-view Hybrid Fusion and Mutual Distillation Network (MV-HFMD) \cite{black2024multi}. We used the authors' implementations and followed the same training and evaluation process as MVPT-NET. The baseline corresponds to our single-view pretrained model, while the single-view output results from MVPT-NET is reported as \(\text{MVPT-NET}_{\text{single-view}}\).

Table \ref{tab1} presents the Benign (Normal) vs. Malignant classification results of MVPT-NET compared to several SOTA models. Our method achieves the highest AUROC of 0.842, outperforming MV-Swin-T \cite{sarker2024mv} which uses the same backbone but suffers from a smaller input resolution and learning from multi-view input from scratch. MV-HFMD \cite{black2024multi} also uses knowledge distillation but faces learning confusion due to its mutual loss leading to the update of the entire model, resulting in a lower AUROC of 0.786. Additionally, the single-view outputs from MVPT-NET also performs better than the single-view pretrained baseline, demonstrating the advantage of our multi-view visual prompt tuning approach in capturing the full range of information from both the MLO and CC views.

\begin{table}
\caption{Benign (Normal) vs. Malignant classification results on the test set. All metrics are reported as the mean$\pm$standard deviation, calculated from the results of 5-fold cross-validation.}\label{tab1}
\begin{tabular}{|l|c|c|c|c|c|}
\hline
Method &  Accuracy	& Precision & Recall & F1-Score & AUROC \\
\hline
MV-Swin-T \cite{sarker2024mv} & 0.683$\pm$0.020 & 0.691$\pm$0.021 & 0.660$\pm$0.015 & 0.676$\pm$0.025 & 0.694$\pm$0.008 \\
\hline
MVC-NET \cite{zhu2021mvc} & 0.728$\pm$0.017 & 0.733$\pm$0.041 & 0.701$\pm$0.030 & 0.711$\pm$0.046 & 0.736$\pm$0.019 \\
\hline
CVT \cite{van2021multi} & 0.708$\pm$0.012 & 0.711$\pm$0.018 & 0.674$\pm$0.024 & 0.704$\pm$0.034 & 0.745$\pm$0.016 \\
\hline
MV-HFMD \cite{black2024multi} & 0.725$\pm$0.028 & 0.708$\pm$0.028 & 0.697$\pm$0.055	& 0.710$\pm$0.065 & 0.786$\pm$0.011 \\
\hline
Baseline (ours) & 0.754$\pm$0.014 & 0.784$\pm$0.030 & 0.743$\pm$0.016 & 0.750$\pm$0.021 & 0.825$\pm$0.007 \\
\hline
MVPT-NET\(_{\text{single-view}} \) & 0.763$\pm$0.007 & 0.744$\pm$0.007 & 0.746$\pm$0.012 & 0.744$\pm$0.007 & 0.837$\pm$0.005 \\
\hline
MVPT-NET\(_{\text{multi-view}} \) & 0.774$\pm$0.006 & 0.757$\pm$0.008 & 0.749$\pm$0.011 & 0.750$\pm$0.007 & 0.842$\pm$0.005 \\
\hline
\end{tabular}
\end{table}

We also conducted a more challenging fine-grained breast cancer classification task, distinguishing between Benign (Normal), Ductal Carcinoma in Situ (DCIS), and Invasive Ductal Carcinoma (Invasive), with the results presented in Table \ref{tab2}. Despite the complexity of the task, our proposed method outperforms all the compared models across all metrics. The success of MVPT-NET can be attributed to pretraining with high-resolution input, avoiding aggressive downsampling, which preserves important details and leads to the learning of a strong backbone. The multi-view visual prompt tuning further learns cross-view features from paired mammogram inputs, updating the prompt without altering the pretrained backbone. This enables the final model to effectively capture both single-view and multi-view information.

\begin{table}
\caption{Benign (Normal) vs. DCIS vs. Invasive classification results on the test set. Precision, recall, F1-score, and AUROC are calculated using the macro average. All metrics are reported as the mean$\pm$standard deviation, calculated from the results of 5-fold cross-validation.}\label{tab2}
\begin{tabular}{|l|c|c|c|c|c|}
\hline
Method &  Accuracy	& Precision & Recall & F1-Score & AUROC \\
\hline
MV-Swin-T \cite{sarker2024mv} & 0.637$\pm$0.023 & 0.607$\pm$0.024 & 0.638$\pm$0.016 & 0.616$\pm$0.028 & 0.701$\pm$0.015 \\
\hline
MVC-NET \cite{zhu2021mvc} & 0.645$\pm$0.019 & 0.612$\pm$0.038 & 0.583$\pm$0.031 & 0.613$\pm$0.041 & 0.730$\pm$0.031 \\
\hline
CVT \cite{van2021multi} & 0.651$\pm$0.014 & 0.604$\pm$0.024 & 0.584$\pm$0.019 & 0.600$\pm$0.022 & 0.754$\pm$0.013 \\
\hline
MV-HFMD \cite{black2024multi} & 0.636$\pm$0.015 & 0.650$\pm$0.047 & 0.629$\pm$0.018 & 0.651$\pm$0.042 & 0.804$\pm$0.010 \\
\hline
Baseline (ours) & 0.668$\pm$0.010 & 0.675$\pm$0.023 & 0.664$\pm$0.012 & 0.664$\pm$0.011 & 0.831$\pm$0.006 \\
\hline
MVPT-NET\(_{\text{single-view}} \) & 0.673$\pm$0.005 & 0.684$\pm$0.007 & 0.666$\pm$0.006 & 0.666$\pm$0.005 & 0.839$\pm$0.005 \\
\hline
MVPT-NET\(_{\text{multi-view}} \) & 0.688$\pm$0.007 & 0.696$\pm$0.005 & 0.681$\pm$0.009 & 0.683$\pm$0.009 & 0.852$\pm$0.004 \\
\hline
\end{tabular}
\end{table}

\subsection{Ablation Study}

To evaluate the contribution of the loss term \( L_{\text{md}} \) in our model, we conducted an ablation study, with results summarized in Table \ref{tab3}. The study was performed on the Benign (Normal) vs. DCIS vs. Invasive classification task. Introducing the \( L_{\text{md}} \) loss term resulted in increase in AUROC, highlighting the effectiveness of knowledge distillation in leveraging meaningful features from multi-view inputs. 

\begin{table}
\caption{Ablation study results on the classification of Benign (Normal) vs. DCIS vs. Invasive. All metrics are calculated from the multi-view outputs. All results are reported as the mean$\pm$standerard deviation, calculated from the results of 5-fold cross validation.}
\label{tab3}
\begin{tabular}{|l|c|c|c|c|c|}
\hline
Method &  Accuracy	& Precision & Recall & F1-Score & AUROC \\
\hline
MVPT-NET w/o \( L_{\text{md}} \) & 0.662$\pm$0.009 & 0.688$\pm$0.012 & 0.655$\pm$0.014 & 0.653$\pm$0.014 & 0.849$\pm$0.003 \\
\hline
MVPT-NET & 0.688$\pm$0.007 & 0.696$\pm$0.005 & 0.681$\pm$0.009 & 0.683$\pm$0.009 & 0.852$\pm$0.002 \\
\hline
\end{tabular}
\end{table}

\section{Conclusions}
In conclusion, we proposed MVPT-NET, a novel approach that adapts Visual Prompt Tuning for multi-view mammogram analysis. Our method outperformed several state-of-the-art multi-view models in breast cancer detection, showcasing its ability to leverage multi-view inputs for more accurate predictions. The ablation study highlights the significance of the introduced loss term in improving model performance by effectively capturing and distilling meaningful features from both MLO and CC views. Our approach, with its high-resolution pretraining and multi-view prompt tuning capabilities, offers a promising direction for enhancing breast cancer detection and classification tasks.

\bibliographystyle{splncs04}
\bibliography{references}

\begin{thebibliography}{10}
\providecommand{\url}[1]{\texttt{#1}}
\providecommand{\urlprefix}{URL }
\providecommand{\doi}[1]{https://doi.org/#1}

\bibitem{black2024multi}
Black, S., Souvenir, R.: Multi-view classification using hybrid fusion and mutual distillation. In: Proceedings of the IEEE/CVF Winter Conference on Applications of Computer Vision. pp. 270--280 (2024)

\bibitem{chen2023teacher}
Chen, H., Jiang, Y., Ko, H., Loew, M.: A teacher--student framework with fourier transform augmentation for covid-19 infection segmentation in ct images. Biomedical Signal Processing and Control  \textbf{79},  104250 (2023)

\bibitem{chen2022unsupervised}
Chen, H., Jiang, Y., Loew, M., Ko, H.: Unsupervised domain adaptation based covid-19 ct infection segmentation network. Applied Intelligence  \textbf{52}(6),  6340--6353 (2022)

\bibitem{chen2024towards}
Chen, H., Martel, A.L.: Towards improved breast cancer detection on digital mammograms using local self-attention-based transformer. In: 17th International Workshop on Breast Imaging (IWBI 2024). vol. 13174, pp. 455--461. SPIE (2024)

\bibitem{chen2022multi}
Chen, Y., Wang, H., Wang, C., Tian, Y., Liu, F., Liu, Y., Elliott, M., McCarthy, D.J., Frazer, H., Carneiro, G.: Multi-view local co-occurrence and global consistency learning improve mammogram classification generalisation. In: International Conference on Medical Image Computing and Computer-Assisted Intervention. pp. 3--13. Springer (2022)

\bibitem{contiero2023causes}
Contiero, P., Boffi, R., Borgini, A., Fabiano, S., Tittarelli, A., Mian, M., Vittadello, F., Epifani, S., Ardizzone, A., Cirilli, C., et~al.: Causes of death in women with breast cancer: a risks and rates study on a population-based cohort. Frontiers in Oncology  \textbf{13},  1270877 (2023)

\bibitem{gou2024research}
Gou, F., Liu, J., Xiao, C., Wu, J.: Research on artificial-intelligence-assisted medicine: a survey on medical artificial intelligence. Diagnostics  \textbf{14}(14), ~1472 (2024)

\bibitem{website}
Halling-Brown, M.D.: Omi-db (2020), \url{https://medphys.royalsurrey.nhs.uk/omidb/}

\bibitem{halling2020optimam}
Halling-Brown, M.D., Warren, L.M., Ward, D., Lewis, E., Mackenzie, A., Wallis, M.G., Wilkinson, L.S., Given-Wilson, R.M., McAvinchey, R., Young, K.C.: Optimam mammography image database: a large-scale resource of mammography images and clinical data. Radiology: Artificial Intelligence  \textbf{3}(1),  e200103 (2020)

\bibitem{he2025dvpt}
He, A., Wu, Y., Wang, Z., Li, T., Fu, H.: Dvpt: Dynamic visual prompt tuning of large pre-trained models for medical image analysis. Neural Networks  \textbf{185},  107168 (2025)

\bibitem{hinton2015distilling}
Hinton, G., Vinyals, O., Dean, J.: Distilling the knowledge in a neural network. arXiv preprint arXiv:1503.02531  (2015)

\bibitem{jain2024follow}
Jain, K., Rangarajan, K., Arora, C.: Follow the radiologist: Clinically relevant multi-view cues for breast cancer detection from mammograms. In: International Conference on Medical Image Computing and Computer-Assisted Intervention. pp. 102--112. Springer (2024)

\bibitem{jia2022visual}
Jia, M., Tang, L., Chen, B.C., Cardie, C., Belongie, S., Hariharan, B., Lim, S.N.: Visual prompt tuning. In: European conference on computer vision. pp. 709--727. Springer (2022)

\bibitem{krizhevsky2017imagenet}
Krizhevsky, A., Sutskever, I., Hinton, G.E.: Imagenet classification with deep convolutional neural networks. Communications of the ACM  \textbf{60}(6),  84--90 (2017)

\bibitem{li2020multi}
Li, C., Xu, J., Liu, Q., Zhou, Y., Mou, L., Pu, Z., Xia, Y., Zheng, H., Wang, S.: Multi-view mammographic density classification by dilated and attention-guided residual learning. IEEE/ACM transactions on computational biology and bioinformatics  \textbf{18}(3),  1003--1013 (2020)

\bibitem{liu2021swin}
Liu, Z., Lin, Y., Cao, Y., Hu, H., Wei, Y., Zhang, Z., Lin, S., Guo, B.: Swin transformer: Hierarchical vision transformer using shifted windows. In: Proceedings of the IEEE/CVF international conference on computer vision. pp. 10012--10022 (2021)

\bibitem{russakovsky2015imagenet}
Russakovsky, O., Deng, J., Su, H., Krause, J., Satheesh, S., Ma, S., Huang, Z., Karpathy, A., Khosla, A., Bernstein, M., et~al.: Imagenet large scale visual recognition challenge. International journal of computer vision  \textbf{115},  211--252 (2015)

\bibitem{sahoo2024systematic}
Sahoo, P., Singh, A.K., Saha, S., Jain, V., Mondal, S., Chadha, A.: A systematic survey of prompt engineering in large language models: Techniques and applications. arXiv preprint arXiv:2402.07927  (2024)

\bibitem{sarker2024mv}
Sarker, S., Sarker, P., Bebis, G., Tavakkoli, A.: Mv-swin-t: mammogram classification with multi-view swin transformer. In: 2024 IEEE International Symposium on Biomedical Imaging (ISBI). pp.~1--5. IEEE (2024)

\bibitem{van2021multi}
Van~Tulder, G., Tong, Y., Marchiori, E.: Multi-view analysis of unregistered medical images using cross-view transformers. In: Medical Image Computing and Computer Assisted Intervention--MICCAI 2021: 24th International Conference, Strasbourg, France, September 27--October 1, 2021, Proceedings, Part III 24. pp. 104--113. Springer (2021)

\bibitem{vaswani2017attention}
Vaswani, A., Shazeer, N., Parmar, N., Uszkoreit, J., Jones, L., Gomez, A.N., Kaiser, {\L}., Polosukhin, I.: Attention is all you need. Advances in neural information processing systems  \textbf{30} (2017)

\bibitem{zhu2021mvc}
Zhu, X., Feng, Q.: Mvc-net: Multi-view chest radiograph classification network with deep fusion. In: 2021 IEEE 18th International Symposium on Biomedical Imaging (ISBI). pp. 554--558. IEEE (2021)

\bibitem{zu2024embedded}
Zu, W., Xie, S., Zhao, Q., Li, G., Ma, L.: Embedded prompt tuning: Towards enhanced calibration of pretrained models for medical images. Medical Image Analysis  \textbf{97},  103258 (2024)

\end{thebibliography}
%






\end{document}